# A REVIEW ON CLOUD ROBOTICS BASED FRAMEWORKS TO SOLVE SIMULTANEOUS LOCALIZATION AND MAPPING (SLAM) PROBLEM

[1]RAJESH DORIYA, [2]PARESH SAO, [3]VINIT PAYAL, [4]VIBHAV ANAND, [5]PAVAN CHAKRABORTY

[1,2,3]National Institute of Technology, Raipur, India-492010
[4]Motilal Nehru National Institute of Technology, Allahabad, India-201004
[5]Indian Institute of Information Technology, Allahabad, India-211012
E-mail: 1rajeshdoriya.it@nitrr.ac.in, 2paresh.h.sao@gmail.com, 3vinitpayal@gmail.com, 4blackquest@gmail.com,
[5]pavan@iiita.ac.in

**Abstract-** Cloud Robotics is one of the emerging area of robotics. It has created a lot of attention due to its direct practical implications on Robotics. In Cloud Robotics, the concept of cloud computing is used to offload computational extensive jobs of the robots to the cloud. Apart from this, additional functionalities can also be offered on run to the robots on demand. Simultaneous Localization and Mapping (SLAM) is one of the computational intensive algorithm in robotics used by robots for navigation and map building in an unknown environment. Several Cloud based frameworks are proposed specifically to address the problem of SLAM, DAvinCi, Rapyuta and C2TAM are some of those framework. In this paper, we presented a detailed review of all these framework implementation for SLAM problem.

## I. INTRODUCTION

Impact of robotic services in different areas are growing, now it is not only limited to laboratory work but it is also being used in robotics services to support daily human activities and now has become one of the emerging research topic. Service robots are mainly designed to perform 4-D operations i.e. dull, dirty, distant and dangerous. Each robotic service has three basic units, sensory unit: to sense the environment, actuation unit: to control mechanical motion and control unit: to overall control of robot.
Each unit give thousands of data in a particular moment. Every year many new sensing device and actuation device are developed and new methods and algorithm are being proposed for efficiency but the real problem arises in handling of that hugely collected data. Apart from this problem, some of the robotic task are very computational expensive and hence cannot be tackled efficiently with on-board robotic hardware.

These problems can be addressed with the use of cloud computing. According to National Institute of Standards and Technology (NIST) Cloud is defined as "a model for enabling ubiquitous, convenient, on demand network access to a shared pool of configurable resources (e.g., servers, storage, networks, applications, and services) that can be rapidly provisioned and released with minimal management effort or service provider interaction".
The functionality of robotics and its computation and information sharing capability could be enhanced by cloud robotics. Cloud robotics is the combination of an ad-hoc cloud formed by machine-to-machine (M2M) communications (participating robots) and an infrastructure cloud enabled by machine-to-cloud (M2C) communications. It utilizes computing model in which resources are dynamically allocated from a shared resource pool in the ubiquitous cloud for offloading and information sharing in robotic applications. One of the most emerging robotics application is Simultaneous Localization and Mapping (SLAM) problem, in this problem robot maps the environment using own position in the environment, during the mapping process it takes thousands of laser and odometry data that needs to be processed efficiently and then the role of cloud computing come to play. There are lot of research papers published on this area, they all agreed to solve this problem with the use of cloud computing. Riazuelo et al. proposed a framework named as C2TAM [9] for cooperative tracking and mapping is achieved with the use of cloud where the computationally expensive tasks such as map optimization and fusing of common area of maps are handled with cloud. It manages the bandwidth required to communicate between robot and cloud server as low as possible. But this framework failed to make the use of independent nature of the data that can give rise to use of parallel and scalable architecture. This is well achieved in DAvinCi framework, this framework was providing scalability and parallelism advantage with cloud for mobile robot nodes. In this framework, parallelism is achieved with Hadoop map reduce computing cluster. But this framework is not made open source hence researchers could make the most of it. Recently a new open source framework Rapyuta is presented by G Mohanarajah et al., it offloads heavy computation by providing secured customizable computing environments in the cloud and also facilitates robots to easily access the RoboEarth Knowledge repository. Each framework has own benefits and drawbacks, during this paper we tried to compare their pros and cons according to their working principles.

## II. SIMULTANEOUS LOCALIZATION AND MAPPING

Estimating location of Landmarks and location of mobile robot simultaneously during map building, is called SLAM: Simultaneous localization and mapping, problem which have highly influence in mobile robot navigation and that is really a hard problem because it have to build a map and at the same moment it has to calculate own position with this map. Solution to this problem enables mobile robots in an unknown environment, to build a consistent map. There is a wide usage of SLAM, like space mobile robot application, underwater application, and outdoor & indoor robot application.

### A. Formulation of SLAM
SLAM problem is considered as two phase problem, first is localization phase and other is mapping phase-
i. Localization: in which robot localize own position on the map which it build simultaneously
ii. Mapping: in this phase it map environment by using laser data and own position. For a mobile robot which is paced on an unknown environment, taking relative observation of a landmarks using a sensor located on the robot at the particular instant we can observe that:

- Robot's controls: an odometry of robot.
  $C_{1:T} = \{c_0, c_1, c_2 \ldots \ldots \ldots c_t\}$
- Observations: a laser scan or camera images of the environment.
  $O_{1:T} = \{o_0, o_1, o_2 \ldots \ldots \ldots, o_t\}$

Odometry and laser scan data is processed and that build the map M of the environment, where the path of the robot can be formulated as

$$X_{0:T} = \{x_0, x_1, x_2 \ldots \ldots \ldots x_t\}$$

There are lot of uncertainties in odometry and the laser scan data, therefore, a probabilistic model is used to model SLAM, so in probability distribution SLAM could be defined as

$$P(X_{0:T}, M \mid O_{1:T}, C_{1:T})$$

Applying Baye's rule gives a framework for sequentially updating the location posteriors, given a map and a transition function

$$p(x_{0:t}, m \mid o_{1:t}, c_{1:t-1}) = p(x_{0:t} \mid o_{1:t}, c_{0:t-1}) \cdot p(m \mid x_{0:t}, o_{1:t})$$

Similarly the map can be updated sequentially by

$$p(m_t \mid x_t, o_{1:t}) = \sum \sum p(m_t \mid x_t, m_{t-1}, o_t) p(m_{t-1}, x_t \mid o_{1:t-1}, m_{t-1})$$

### B. Solution to SLAM Problem
There are large varieties of different SLAM approaches have been proposed, but the underlying concept is almost same in each, consisted with five basic steps.

1. Landmark Extraction: Process of features collection by observing the environment is called Landmark which have some properties like,

a. Landmarks should be re-observable.
b. Landmarks can be identified uniquely.
c. Environment should have sufficient number of landmarks.
d. Landmarks should be at fixed position.

Robot sensory unit measure laser scan of the environment and extract features from the environment.
2. Data Association: The process of matching and linking observed landmarks patterns from different other patterns. This can be also referred as re-observing landmarks.
3. State Estimation: The robots then attempt to associate these landmarks to previous one and robot's position is updated using re-observed landmarks in the EKF (Extended Kalman Filter). EKF can estimate current robot state from odometry data and landmark observation.
4. State Update: Any new landmark observed are added as new observations so when the environment is observed later they can be detected.
5. Landmark Update: In the last step of SLAM, when the new landmarks are observed they are added in the robot map of the world to build map more reliable. This is achieved by establishing the relation between old landmarks and new landmarks.

## III. CLOUD COMPUTING

Cloud Computing is a computing metaphor evolved around grid computing and Service Oriented Architecture (SOA). It allows users to access computing resources over the network as per demand. Cloud computing paradigm can be classified as public, private and hybrid.

There are three service delivery models of cloud computing: Software as a Service (SaaS), Platform as a Service (PaaS) and Infrastructure as a Service (IaaS).

However, one can introduce X as a service by incorporating its standard requirement to be a service. Now a days cloud computing is being used in various domains and applications. Various researchers have explored its potential of utilization in the field of Robotics.

It improves performance of robot by offloading its computational tasks and can also store the data.

## IV. CLOUD COMPUTING BASED FRAMEWORKS FOR SLAM

There are number of frameworks suggested by various researchers. Based on literature DAvinCi, Rapyuta and C2TAM frameworks have got wide attention in the robotics research people:

### A. DAvinCi Framework

This framework is one of the very popular framework in cloud robotics. In this framework, ROS packages are used to provide abstraction over the robotic hardware. Robot passes its data over HTTP to the cloud server which in turn passes to the Hadoop map reduce computing cluster. The laser scan data of robots is processed with Hadoop map reduce computing cluster. In the implementation, eight nodes are used in map reduce cluster and results in a significant performance over a single node. The performance of the system can be further enhance by incorporating more number of computing nodes. Hadoop map reduce computing cluster uses Hadoop Distributed File System (HDFS) to handle the data storage. All robots will share a global map in that environment.

The working of the DAvinCi framework is shown in the Figure 1. Architecture of DAvinCi framework In DAvinCi framework, ROS is used to provide a communication and messaging system, between robots and DAvinCi server. Any ROS message, which is being send to DAvinCi server is wrapped in HTTP requrest and on the server side the message is processed after unwrapping and same steps are performed in reverse order at the robot's hand to get the response. When Hadoop map reduce computing cluster gets the data, it divides the data into chunks of equal sizes and process it with computing nodes. Main components of DAvinCi architecture are described below-

#### DAvinCi Server

DAvinCi server serves as a mediator to provide services to robots. With the help of ROS it provides communication ecosystem in the overall architecture.

#### HDFS cluster

The HDFS cluster consists of various computational nodes. Each node perform two basic operations, map and reduce. Map operation is used to operate data with same key value pair and reduce operation summarized the results obtained from map operations.

These nodes are managed by Hadoop Distributed File System (HDFS). This enable nodes to work in parallel across the cluster in form of map and reduce operations. The HDFS cluster also provide in-built fault tolerance facility.

#### Mobile robot Node

A mobile robot is used to perceive the environment. It can communicate with server. So that it can send it's captured key frames to server and receive a response from server and use that response to take decision.

### B. Rapyuta Framework

This is an open source cloud platform designed specifically for robots, it is a type of Platform as a Service (PaaS) cloud delivery model consisting of Operating System, executing platform, a database and communication server.

It provides a user friendly interface to offload robotic data and also facilitates computing environment to access knowledge repository of RoboEarth project.

Architecture of Rapyuta is simple but does not compromise with security, each robot in Rapyuta framework gets a secured computing environment, that computing environment are tightly interconnected with each other which allowed robots to share services and data that provides an excellent platform for multi-robot deployments.

In this framework, computing environments are connected to RoboEarth knowledge repository that provides different datasets of different Robots and this could lead the robots to gain experience from other robots.

The Rapyuta's Web Socket-based communication server provides bidirectional, full duplex communications with the physical robot. Figure 2 shows the end-to-end point connectivity over RPC of Rapyuta framework.

Following are the basic components of Rapyuta framework:
a. Computing Environment computing environment is implemented using Linux containers.
Along with the isolation of processes and resources within a single host with native speed, it also provides different other functionalities such as configuration of memory limits, disk quotas, I/O limits, CPU quota etc.

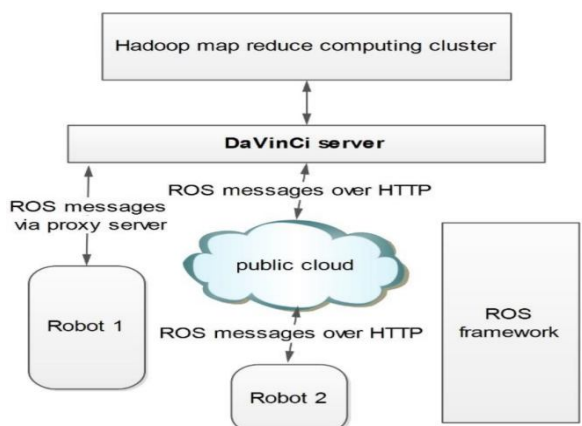

Figure 1: DAvinCi framework [10]

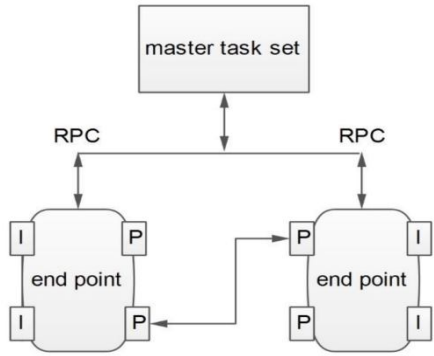

Figure 2: Rapyuta framework [12]

b. Communication Protocols

Rapyuta's communication protocols have different Endpoints which represents a process, all endpoints are connected by RPC and have interface and ports. Interface are used for external communication e.g. non-Rapyuta processes running on the robot or in the computing environment and the socket ports are utilized internal communication between endpoint process.

c. Core tasks

Basically it comprises of four tasks:

i. Master: The main job of master is to monitor and maintain the command data structure between the robots and Rapyuta. It also monitors network of other task sets.

ii. Robot: It covers communication between robot and endpoint, forwarding message to master and conversion of data messages.

iii. Environment: It facilitates the communication with ROS nodes and other endpoints. It also starts and stops ROS nodes, adds and removes parameters.

iv. Container: It provides capability necessary steps to start and stop computing environment.

d. Command Data Structure It has four components:

i. Network: It is the core part of whole platform which provides abstraction and referred as command data structure. It also manages internal and external communication system.

ii. User: An API key is the unique key used for robot's authentication.

iii. Load Balancer: It is used for managing machines which are used in computing environment and also used to assign new machine in appropriate computing environment.

iv. Distributer: It distributes the incoming connections from the robots to other available online robot end points.

C. C2TAM Framework

C2TAM stands for Cloud Framework for cooperative tracking and mapping. In this framework, computational extensive algorithms are offloaded to cloud such as map optimization and fusing of common area of maps. This framework is able to keep the bandwidth requirement on the lower side for communication between robot and cloud server. When some frames are sent to cloud server, it checks for any overlapping area with any existing map and if found, it fuses new map with that map.

Table 1: A comparision table of frameworks for SLAM

| Parameters | C2TAM framework | Rapyuta framework | DAvinCi framework |
|---|---|---|---|
| Key Technology | It uses multithreading. | A Linux containers is used to isolate processes and resources within a single host with native speed. | It uses Hadoop map reduce computing cluster and parallelize the fast SLAM algorithm. |
| Communication between server and robot | It uses ROS framework for communication. | It provide ROS framework and a web Socket-based framework for communication. | In DAvinCi framework ROS was used to provide a communication and messaging services to robots and DAvinci server. |
| Communication bandwidth | It works on a particular thread at a time hence low bandwidth is required. | Process size are bigger then comparing to thread so high bandwidth are required. | Requires high bandwidth utilization due to the use of Hadoop map reduce computing cluster. |
| Computational latency | It allocate the expensive map optimization process out of the robot platform that allows a significant reduction in the on-board computation. | Comparatively encompasses high computational latency. | Use of parallelization results in low computational latency. |
| Power consumption | Low power consumption. | High power consumption. | Low power consumption. |
| Map building approach | It uses centralized map building approach that allows a straightforward massive data storage of robotic sequences and geometric estimations. | It uses elastic model and dynamic map building approach. | Centralized map building approach is used |
| Additional feature | It provides interface to an advanced database of visual maps in the Cloud. | It provides very user friendly and customizable environment to offload huge computational data. | It provides highly map optimization feature. |

In this framework, parallel tracking and mapping is used to make it more efficient. Two parallel threads run simultaneously one thread produces 3D map by using bundle adjustment. And second thread tracks and estimates the camera location using a known map. For optimization, tracking process does not sends key frames continuously it only sends key frames containing new information. In this way it saves bandwidth. The working of C2TAM framework is shown in Figure 3.

The following SLAM components are used in C2TAM framework- Mapping
It uses sensor data to estimate model of scene. In optimized implementation two threads are used in which first thread estimates position of camera and second thread performs task of map estimation. And this optimization has shown its advantage in form of low computational latency and it also does not harm the accuracy.

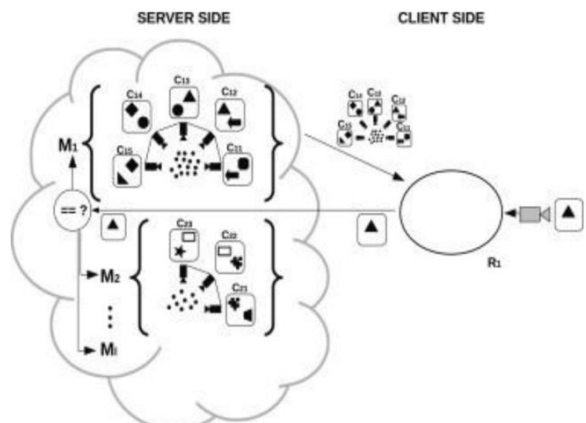

Figure 3: C2TAM framework [9]

Tracking
Tracking component is responsible for estimating position of camera after certain time period assuming time is t in a known map M. All nodes have a tracking component say there are n components then there exist n tracking components which track camera estimation of their respective nodes.

Relocation
Relocation component comes in light only if any reason mapping component fails the reasons due to which tracking component can fails and results in suddenly high acceleration, and fails in interpreting the key frames captured by camera. In this case, relocation component relocates camera and re-start tracking and robot continues to work.

Place Recognition
Place recognition is capability of relocating in large numbers of maps. This is also called kidnapped robot problem. It is slightly different from relocation because relocation is used when we suddenly lost the previously tracked location. The tracking position cannot be determined just before failure occurred and this position location is used to find current location having that its current location is somewhere around the last known location but in this place recognition actual location is not known. This can be find by comparing its key frames with all available maps and find the map in which this location belongs to. Place recognition is comparatively more computation extensive process because all locations are equally likely.

Map fusion
It is the process of merging two maps which are having overlapped area. This will be required when we are having two robots which are capturing the same area.

Table 1 presents the comparative study of all the discussed frameworks for SLAM problem.

**CONCLUSION**

In this paper, we discussed about different cloud frameworks and compared how they interconnected and differ in their working principles for SLAM problem. Each framework used different set of protocols which allow robots offload the computation to the cloud. On the basis of observed facts, we believe that Rapyuta framework is more flexible framework than the other. Due to the open source implementation of the Rapyuta framework, it can be extended to other robots functionalities easily by other researchers. Whereas, when the high degree of parallelism is desirable DAvinCi architecture is best suited for SLAM. C2TAM framework is found suitable in the condition of huge data storage and low bandwidth availability.

★ ★ ★